\documentclass{article}

 \PassOptionsToPackage{numbers, compress}{natbib}

\usepackage[final]{nips_2018}

\usepackage[utf8]{inputenc} 
\usepackage[T1]{fontenc}    
\usepackage[colorlinks]{hyperref}       
\usepackage{url}            
\usepackage{booktabs}       
\usepackage{amsfonts}       
\usepackage{nicefrac}       
\usepackage{microtype}      
\usepackage{graphicx}
\usepackage{float}
\usepackage{comment}
\usepackage{bbm}
\usepackage{textcomp}
\usepackage{color}
\usepackage{cleveref}

\title{A Multimodal Dialogue System for Conversational Image Editing}

\author{
  Tzu-Hsiang Lin$^1$ \quad  Trung Bui$^2$ \quad Doo Soon Kim$^2$ \quad Jean Oh$^1$ \\
    $^{1}$Carnegie Mellon University, \quad $^{2}$Adobe Research\\
    \{tzuhsial,hyaejino\}@andrew.cmu.edu, \quad \{bui,dkim\}@adobe.com
}

\begin{document}

\maketitle


\begin{abstract}
    In this paper, we present a multimodal dialogue system for Conversational Image Editing.  We formulate our multimodal dialogue system as a Partially Observed Markov Decision Process (POMDP) and trained it with Deep Q-Network (DQN) and a user simulator. Our evaluation shows that the DQN policy outperforms a rule-based baseline policy, achieving 90\% success rate under high error rates.  We also conducted a real user study and analyzed real user behavior.
\end{abstract}

\section{Introduction}

Image editing has been a challenging task due to its steep learning curve. Image editing software such as Adobe Photoshop typically have a complex interface in order to support a variety of operations including selection, crop, or slice. They also require users to learn jargons such as \textit{saturation}, \textit{dodge} and \textit{feather}\footnote{\textit{saturation}: color intensity; \textit{dodge}: tool to lighten areas; \textit{feather}: tool to create soft edges}. In order for a user to achieve a desired effect, some combinations of operations are generally needed. Moreover, image edits are usually localized in a particular region, for instance, users may want to add more color in the trees or remove their eyepuffs. In the first case, users need to first select the trees and then adjust the saturation to a certain level.  In the second case, users need to select the eyepuffs, apply a reconstruction tool that fills the region with nearby pixels, then apply a patching tool to make the reconstruction more realistic. Such complexity makes the editing task challenging even for the experienced users.

In this paper, we propose a conversational image editing system which allows users to specify the desired effects in a natural language and interactively accomplish the goal via multimodal dialogue. We formulate the multimodal dialogue system using the POMDP framework and train the dialog policy using Deep Q-Network (DQN)\cite{mnih2015human}. To train the model, we developed a user simulator which can interact with the system through a simulated multimodal dialogue. We evaluate our approach--i.e., DQN trained with the user simulator--by comparing it against a hand-crafted policy under different semantic error rates. The evaluation result shows that the policy learned through DQN and our user simulator significantly outperforms the hand-crafted policy especially under a high semantic error rate. We also conducted a user study to see how real users would interact with our system.

The contributions of the paper are summarized as follows:
\begin{itemize}
    \item We present a POMDP formulated multimodal dialogue system.
    \item We developed a multimodal multi-goal user simulator for our dialogue system.
    \item We present an architecture for Conversational Image Editing, a real-life application of the proposed framework in the domain of image editing  
    \item We present the experiment results of comparing the proposed model against a rule-based baseline.
\end{itemize}

\section{Related Work}
The multimodal system PixelTone~\cite{laput2013pixeltone} shows that an interface combining speech and gestures can help users increase more image operations. Building on its success, we propose to build a multimodal image editing dialogue system.  Previous research on multimodal dialogue systems mostly focus on the architectures~\cite{trung2006multimodal} for multimodal fusion and did not adopt the POMDP framework~\cite{young2013pomdp}. Since the dialogue managers of these systems are based on handcrafted rules, it cannot be directly optimized.  Also, real users are essential in evaluating these systems~\cite{whittaker2005evaluating}, which can be costly and time inefficient.

Information-seeking dialogue systems such as ticket-booking~\cite{el2017frames} or restaurant-booking~\cite{dhingra2017towards, wen2017network} typically focus on achieving one user goal throughout an entire dialogue. Trip-booking~\cite{el2017frames} is a more complex domain where memory is needed to compare trips and explore different options. The most similar domain is conversational search and browse~\cite{heck2013multi, hakkani2014eye}, where the system can utilize gestures and even gazes to help users to locate the desired objects.  A recently collected corpus~\cite{dial_edit} shows that Conversational Image Editing is more challenging,  requiring to address not only these aspects but also the composite-task setting ~\cite{Peng2017CompositeTD,Budzianowski2017SubdomainMF} where the user may have multiple goals to achieve in a sequential order.

Our task is also related to Visual Dialogue~\cite{visdial} which focuses on the vision and the dialogue jointly. The agent in Visual Dialogue needs to recognize objects, infer relationships, understand the scene and dialogue context to answer questions about the image.  Our agent also requires a similar level of understanding, but the focus on vision is more of recognizing a localized region in an image.  Another closely related area is vision-and-language navigation~\cite{anderson2018vision}, since both the navigation instructions (e.g., go upstairs and turn right) and image edit requests~\cite{editme} (e.g., remove the elephant not looking forward) are mostly in imperative form and relates to what the agent sees.

\section{Partially Observed Markov Decision Process Formulation} \label{sec:pomdp}

In this section, we formulate our image editing dialogue system as a Partially Observable Markov Decision Process (POMDP)~\cite{young2013pomdp}.  POMDP dialogue framework combines \textit{Belief State Tracking} and \textit{Reinforcement Learning}.  \textit{Belief State Tracking} represents uncertainty in dialogue that may come from speech recognition errors and possible dialogue paths, whereas \textit{Reinforcement Learning} helps the dialogue manager discover an optimal policy.

POMDP is composed of belief states $B$, actions $A$, rewards $R$, and transitions $T$.  The goal is to optimize a parametrised policy $\pi:B\rightarrow A$ to maximize the expected sum of rewards $R=\sum_{t}\gamma^t r_t$. 

\subsubsection*{State}

Our state space $B=B^u\oplus B^e$ includes the user state $B^u$ and the image edit engine state $B^e$. $B^u$, the estimation of the user goal at every step of the dialogue, is modeled as the probability distribution over possible slot values.  For gesture related slots (e.g., \textit{gesture\_click}, \textit{object\_mask\_str}), we assume these values hold 100\% confidence and assigns probability score $1$ if a value is present and $0$ otherwise. $B^e$ is the information provided by the engine that is related to the task at hand.  

The main difference from convention dialogue systems is that we include information from the engine.  Since the image edit engine displays the image, executes edits and stores edit history, we hypothesize that including this information can help our system achieve a better policy.  One example is, if the edit history is empty, users will unlikely to request an \textit{Undo}. Examples of our state features is presented in Table \ref{tab:state}.

\begin{table}[H]
    \centering
    \begin{tabular}{c|c|c}
      Type  &  Feature Type & Examples  \\
     \hline 
      Speech    &  Distribution over possible values & \textit{intent, attribute} \\
      Gestures  &  Binary  & \textit{image\_path} \\
      Image edit engine  &  Binary  & \textit{has\_next\_history} \\
    \end{tabular}
    \vspace{5pt}
    \caption{Example state features used for dialogue policy learning}
    \label{tab:state}
\end{table}

\subsubsection*{Action}
We designed 4 actions for our dialogue system: (i) \textit{Request}, (ii) \textit{Confirm}, (iii) \textit{Query}, and (iv) \textit{Execute}.  \textit{Request} and \textit{Confirm} actions are each paired with a slot. \textit{Request} asks users for the value of a slot and \textit{Confirm} asks users whether the current value stored by the system is correct. \textit{Query} takes the current value in slot \textit{object} and queries the vision engine to predict segmentation masks of \textit{object}.
\textit{Execute} is paired with an intent. \textit{Execute} passes its paired intent and the intent's children slots (\Cref{fig:ontology}) to the image edit engine for execution.  If any of the arguments are missing, the execution will fail, and the image will remain unchanged.

Unlike information-seeking dialogue systems~\cite{dhingra2017towards, wen2017network} which attempt to query a database at every turn, we make \textit{Query} an independent action because of two reasons: (i) modern vision engines are mostly based on Convolutional Neural Networks (CNNs) and frequent queries may introduce performance latency; (ii) segmentation results should be stored, and consecutive queries will override previous results. 

\subsubsection*{Reward}
We define two reward functions.  The first reward function is defined based on PyDial~\cite{ultes2017pydial}.  The reward is given at the end of a dialogue and defined as $20 * \mathbbm{1}(D) - T $, where $20$ is the success reward, $\mathbbm{1}(D)$ is the success indicator and $T$ is the length of the dialogue.  The second reward function gives a positive reward $r^p$ when an user goal is completed, and a negative reward $r^n$ if an incorrect edit is executed.  The main idea for the second reward function is that since image editing dialogues have multiple image edit requests, additional supervision reward will better help train the dialogue system.  However, we did not observe a huge difference between the two reward functions in our initial experiments. Therefore, we only present the results on the first reward function.

\subsubsection*{Transition}  Our transitions are based on the user simulator and the image edit engine.  Every time step $t$, the system observes belief state $b_t$ and outputs system action $a_t$. The image edit engine observes system action $a_t$ and update its state to $b^e_{t+1}$. The user simulator then observes both $b^e_t$ and $a_t$ then updates it state to $b^u_{t+1}$. Next state $b_t=b^u_t\oplus b^e_t$ will pass to the system for the next turn.  Both the user simulator and the image edit engine are updated according to predefined rules.

\subsection*{Dialogue Policy}

We present two polices for dialogue management.

\textbf{Rule-based}: We hand-crafted a rule-based policy to serve as our baseline.  The rule-based policy first requests the intent from the user.  After knowing the intent, it then requests all the slots that correspond to that particular intent and then executes the intent. To obtain the localized region (\textit{object\_mask\_str}), the rule-based policy first  queries the vision engine and then requests \textit{object\_mask\_str} if the vision engine result is incorrect.

\textbf{Deep Q-Network}: Deep Q-Network~\cite{mnih2015human} combines artificial neural networks and reinforcement learning and takes state $b^t$ as inputs to approximate action values $Q(s_t,a_t)$ for all action $a$. Deep Q-Networks are shown to succeed in spoken dialogue systems learning~\cite{ultes2017pydial} due to its capability to model uncertainty in spoken language understanding and large domain space.

\section{Conversational Image Editing System}

In this section, we first present the ontology used in our system, then describe the role of each system component in further detail.

\subsection{Domain Ontology}
\label{subsec:ontology}

\begin{figure}
    \centering
    \includegraphics[width=0.8\textwidth]{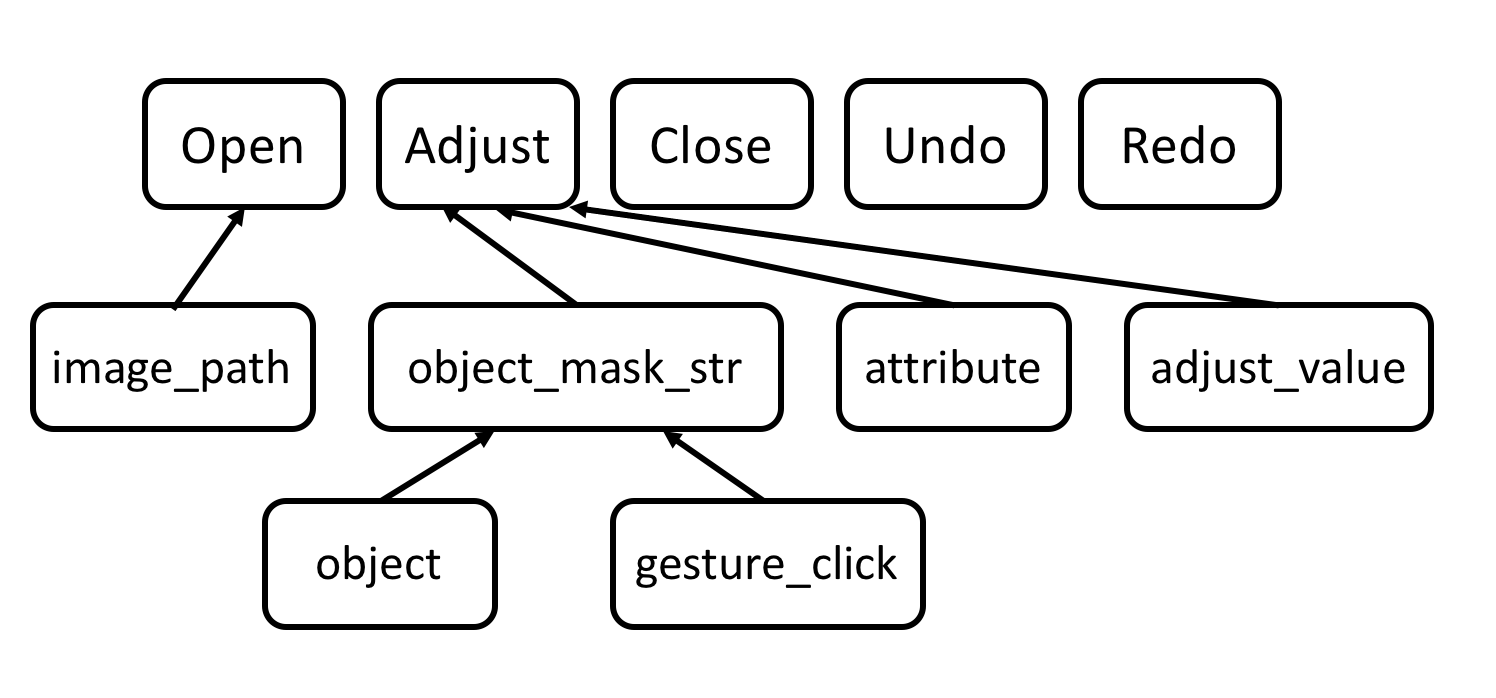}
    \caption{This figure depicts the domain ontology of our conversational image editing system.  Top level nodes represent intents; Mid level and low level nodes represent slots. Arrows indicate dependencies.  Mid level nodes that directly point to top level nodes are arguments directly associated with that intent.  Right three intents do not require additional arguments. }
    \label{fig:ontology}
\end{figure}

Conversational Image Editing is a multimodal dialogue domain that consists of multiple user intents and a natural hierarchy.  Intents are high level image edit actions that may come with arguments or entities~\cite{dial_edit, manuvinakurike2018conversational}.  Also, most edit targets are localized regions in the image.  This introduces a hierarchy where the system has to first request the name of the object and then query vision engine to obtain the object's segmentation mask.

We handpicked a subset of user intents from \textit{DialEdit}~\cite{dial_edit} and \textit{Editme}~\cite{editme} corpus.  The user intents are \textit{Open, Adjust, Close, Undo, Redo}. 
\textit{Open} and \textit{Close} are inspired by ~\citet{manuvinakurike2018conversational}; 
\textit{Adjust} modifies the attributes of a region; \textit{Undo} reverts incorrect edits and \textit{Redo} can redo edits if \textit{Undo} is accidentally executed.
\textit{Open} requires slot \textit{image\_path}; \textit{Adjust} requires slots \textit{object\_mask\_str, adjust\_value, and attribute}.  Slot \textit{object\_mask\_str} further depends on slots \textit{object}, and \textit{gesture\_click}. \textit{Close, Undo}, and \textit{Redo} do not require slots. Our ontology is depicted in \Cref{fig:ontology}.

\subsection{Components}

\begin{figure}
    \centering
    \includegraphics[width=0.8\textwidth]{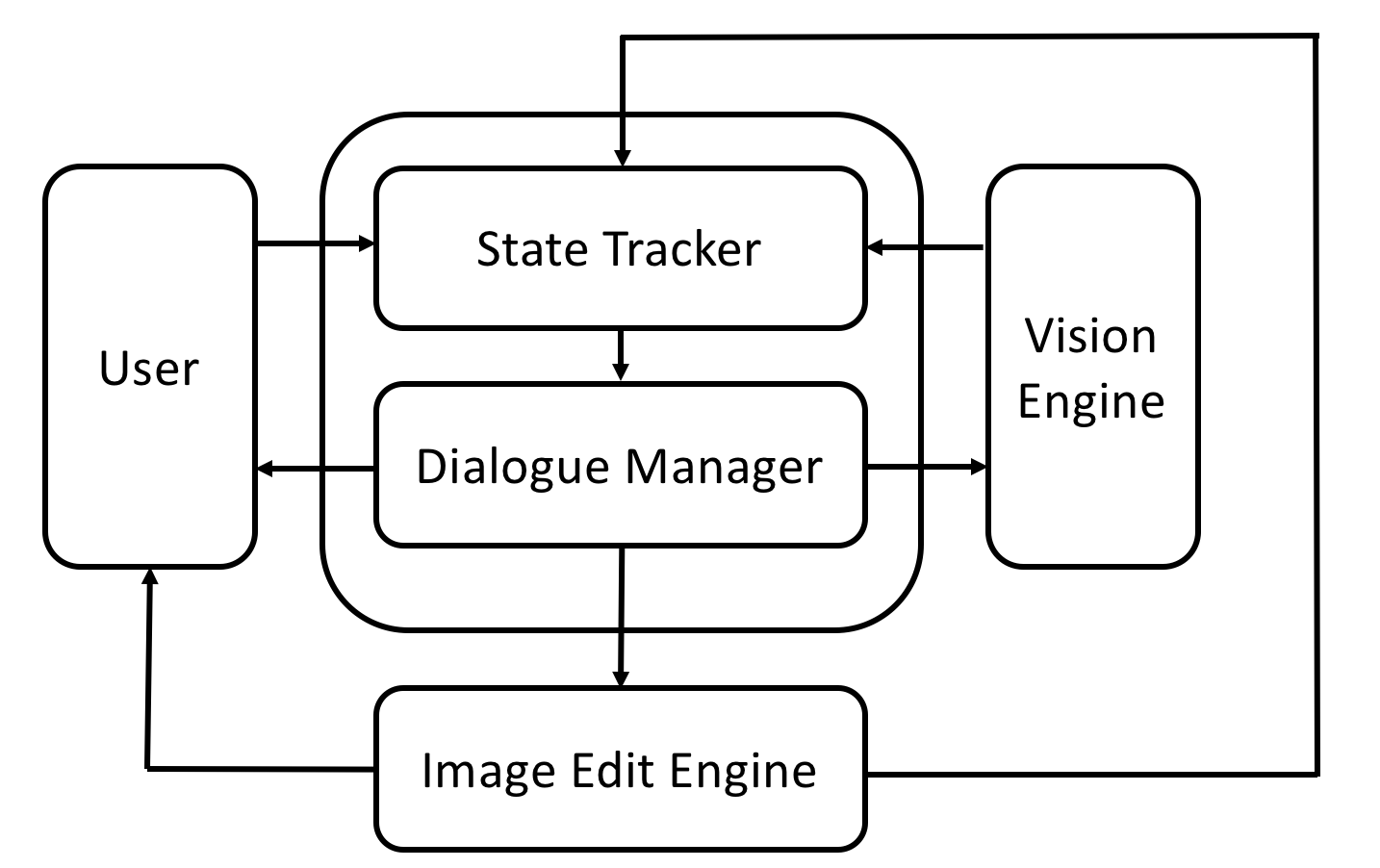}
    \caption{This figure illustrates the components in our system and interactions with the user. State Tracker observes information from User and Image Edit Engine then passes the state to Dialogue Manager.  Dialogue Manager then selects an action.  If the action is \textit{Query}, Dialogue Manager will query Vision Engine and store the results in the state.  The action will then be observed by both the User and Image Edit Engine. 
    }
    \label{fig:dialoguesystem}
\end{figure}

Our system architecture consists of four components: (i) Multimodal State Tracker (ii) Dialogue Manager (iii) Vision Engine (iv) Image Edit Engine

\subsubsection*{Multimodal State Tracker}
Input to our state tracker consists of three modalities (i) user utterance, (ii) gestures (iii) image edit engine state. For (i), we trained a two-layer bi-LSTM on the \textit{Editme}~\cite{editme} corpus for joint intent  prediction and entity detection.  We select the \textit{Editme} corpus because it is currently the only available dataset which contains image editing utterances.  \textit{Editme} has about 9k utterances and 18 different intents.  Since \textit{Editme} is collected by crowd-sourcing image edit descriptions, the utterances are often at a high level, and the annotation can span up to a dozen words. For example, in \textit{Editme}, "lower the glare in the back room wall that is extending into the doorway" has the following labels: intent is labeled as "adjust"; attribute is labeled as "the glare"; region is labeled as "in the back room wall that is extending into the doorway".  Our tagger achieved 0.80\% intent accuracy and 58.4\% F1 score on the validation set.
Since there exists a discrepancy between our ontology and \textit{Editme}, an additional string matching module is added to detect numerical values and intents not present in \textit{Editme}.  For (ii), we directly take the output as state values.  That is, if gestures are present, then the gestures slot values will 1.  For (iii), we designed several features including ones mentioned in Table \Cref{tab:state}.  The final output is the concatenation of the results of (i), (ii) and (iii), which is the belief state $B$ in the previous section.


\subsubsection*{Dialogue Manager}

Dialogue manager is a rule-based policy or a parametrised policy that observes the dialogue state, and performs actions to complete user's image edit requests.  Detailed of actions are presented in the previous section.

\subsubsection*{Vision Engine} 

Vision engine performs semantic segmentation and is called when system selects action \textit{Query}.  Vision engine takes the image and slot \textit{object}'s value as query and outputs a list of candidate masks to be shown to the user.  If present, \textit{gesture\_click} will be used filter the candidate masks by checking whether it overlaps with any of the candidate masks.  We leave extracting state features from vision engine as future work.
    
\subsubsection*{Image Edit Engine} 

Image edit engine is an image edit application that acts as an interface to our user and as an API to our system.  At every turn, our system loads the candidate masks stored in slot \textit{object\_mask\_str} to the engine for display.  When system performs an \textit{Execute} action, the executed intent and associated arguments will be passed to the engine for execution.  If the intent and slots are valid, then an edit can be performed. Else, the execution will result in failure, and image will remain unchanged. 

We developed a basic image edit engine using the open-source OpenCV~\cite{opencv_library} library.  The main features of our engine include image edit actions, region selectors, and history recording.  

\subsection{Multimodal Multi-goal User Simulator}

We developed an agenda based~\cite{schatzmann2007agenda} multimodal multi-goal user simulator to train our dialogue system. User agenda is a sequence of goals which needs to be executed in order.  A goal is defined as an image edit request that is composed of an intent and it depending slots in the tree~\Cref{fig:ontology}. A successfully executed goal will be removed from the agenda, and the dialogue is considered as success when all the user goals are executed. If an incorrect intent is executed, the simulator will add an \textit{Undo} goal to the agenda to undo the incorrect edit.  On the other hand, if the system incorrectly executes \textit{Undo}, the simulator will add a \textit{Redo} goal to the agenda.  Our simulator is programmed not to inform \textit{object\_mask\_str} unless asked to.

\section{Simulated User Evaluation}

\subsection{Experimental setting}

\textbf{User Goal}  We sampled 130 images from MSCOCO~\cite{lin2014microsoft} and randomly split them into 100 and 30 for training and testing, respectively.  Goals are randomly generated using our ontology and the dataset.  For simulated users, we set the number of goals to 3 which starts with an \textit{Open} goal, followed by an \textit{Adjust} goal, then ends with a \textit{Close} goal.  Maximum dialogue length is set to 20.  To simulate novice behavior, a $\theta$ parameter is defined as the probability a slot will be dropped in the first pass. We set $\theta$ to 0.5. 

\textbf{Vision Engine}  We directly take ground truth segmentation masks from the dataset as query results.

\textbf{Training Details} For DQN, we set the hidden size to 40.  We set batch size to 32 and freeze rate to 100. We used 0.99 for reward decay factor $\gamma$.  The size of experience replay pool is 2000. We used Adam~\cite{kingma2014adam} and set the learning rate to 1e-3. 

\subsection{Results}

\begin{table}[]
    \centering
    \begin{tabular}{c|c c c c |c c c c}
    & \multicolumn{4}{c}{Rule-based} &\multicolumn{4}{c}{DQN} \\
    SER & Turn & Reward & Goal & Success  & Turn & Reward & Goal & Success    \\ 
    \hline 
    \hline 
    0.0 & 7.5 & 13.50 & 3.0  & 1.0 & 7.43 & 13.56 & 3.0  & 1.0  \\
    \hline
    0.1 & 7.2 & 13.80 & 3.0  & 1.0 & 7.27 & 13.73 & 3.0  & 1.0  \\
    \hline
    0.2 & 9.16 &  11.13 & 2.9  & 0.97 & 8.33 & 12.66 & 3.0  & 1.0  \\
    \hline
    0.3 & 15.0 & 4.6 & 2.87  & 0.93 & 9.23 & 11.76 & 3.0  & 1.0  \\
    \hline
    0.4 & 17.3 & -6.1 & 2.30  & 0.53 & 12.1 & 8.2 & 2.93  & 0.97  \\
    \hline
    0.5 & 18.2 & -16.8 & 1.50  & 0.07 & 13.6 & 5.3 & 2.8  & 0.90  \\
    \hline
    \end{tabular}
    \vspace{5pt}
    \caption{Results of our simulated user evaluation.  SER denotes semantic error rate.  Our DQN policy still retains high success rate even under high semantic errors compared to rule-based baseline. }
    \label{tab:simresults}
\end{table}

We evaluate four metrics under different semantic error rates (SER): (i) Turn (ii) Reward (iii) Goal (iv) Success.  (i), (ii), and (iv) follow standard task-oriented dialogue system evaluation.  Since image editing may contain multiple requests, having more goals executed should indicate a more successful dialogue, which success rate cannot capture.  Therefore, we also include (iii).  

\Cref{tab:simresults} shows the results of our simulated user evaluation. 
At low SER (<0.2), we can see that rule-based and DQN can successfully complete the dialogue.  As SER increases (>0.2), the success rate of the rule-based policy decreases.  On the other hand, our DQN policy manages to learn a robust policy even under high SERs and achieved 90\% success rate.

\section{Real User Study}

\begin{table}[]
    \centering
    \begin{tabular}{c | c c | c c}
    & \multicolumn{2}{c}{Rule-based} &\multicolumn{2}{c}{DQN} \\
           & Turn & Success & Turn & Success  \\ 
    \hline 
    \hline 
    Real User & 5.7  & 0.8   & 3.9 & 0.8  \\
    \hline
    \end{tabular}
    \vspace{5pt}
    \caption{Result metrics of our real user study on the 10 sampled goals from the test set.  Rule-based and DQN policies have the same 0.8 success rate, and the DQN policy has fewer number of turns.}
    \label{tab:realresults}
\end{table}

While the experiment in the previous section shows effectiveness under a simulated setting, real users may exhibit a completely different behavior in an multimodal dialogue.  Therefore,  we built a web interface(\Cref{sec:interface} for our system.  Our interface allows text input and gestures. Users can input \textit{gesture\_click} by putting a marker on the image, and input \textit{object\_mask\_str} by putting a bounding box on the image as the localized region. 

We recruited 10 subjects from the author's affiliation and conducted a study testing them against our Rule-based Policy and the DQN policy. One person among the 10 is familiar with image editing. 10 goals were sampled from the test set, and each person is asked to complete two dialogues of the same policy. In every dialogue, there is only one \textit{Adjust} goal and it is presented to the user in semantic form.  Since number of goals is reduced, we set maximum dialogue length to 10.

\Cref{tab:realresults} presents the turns and success rate in our real user study. Both policies achieved 0.8 success rate, while the DQN policy has fewer number of turns.  Due to the small amount of goals, we focus on the insights gained from manually inspecting the dialogue data.  

\textit{Turn Number}: We discovered the turn number in our study mostly depend on the user behavior.  For users who prefer to express more information, the turn number is smaller. For users who prefer to interact mostly with text and leave gestures as a last resort, the turn number is larger. 

\textit{Success Rate}:  The semantic form of the user goal makes it easy for the users to express and inform slot values (\Cref{sec:interface})), so that most dialogues are successful.  All the failure cases are incorrect understanding of the slot \textit{adjust\_value}, which is a result of the string matching module in our state tracker. \Cref{tab:successdialogue} presents a success dialogue and \Cref{tab:failuredialogue} presents a failure dialogue.

\begin{table}[]
    \centering
    \begin{tabular}{c|c|l|c}
    Turn  & Speaker & Utterance & Gesture \\
    \hline
    \hline 
    1 & User & increase the man's saturation by 10 \\
    1 & System & What object\_mask\_str do you want? & \\
    2 & User & man & object\_mask\_str \\
    2 & System & Execute: intent=adjust, adjust\_value=10 & \\
      &        & attribute=saturation, object\_mask\_str=iVBOR & \\
    \hline
    \multicolumn{4}{c}{Success}
    \end{tabular}
    \vspace{5pt}
    \caption{Success dialogue example with DQN policy.  The user first interacts only with text input in the first turn, then inputs both the "man" (\textit{object}) and gesture (\textit{object\_mask\_str}) in the second turn. }
    \label{tab:successdialogue}
\end{table}

\begin{table}[]
    \centering
    \begin{tabular}{c|c|l|c}
    Turn  & Speaker & Utterance & Gesture \\
    \hline
    \hline 
    1 & User & make the man {\color{blue}30\% less} bright & object\_mask\_str \\
    1 & System & Execute: intent=adjust, {\color{red}adjust\_value=30} \\
      &        & attribute=brightness, object\_mask\_str=iVBOR & \\
    \hline
    \multicolumn{4}{c}{Failure}
    \end{tabular}
    \vspace{5pt}
    \caption{Failure dialogue example with DQN policy.  In the first turn, our issues a image edit request and also the localized region (\textit{object\_mask\_str}).
    However, our state tracker failed to interpret "30\% less bright" into the value "-30", and executes with the incorrect value 30 (\textit{adjust\_value}).
    }
    \label{tab:failuredialogue}
\end{table}

\section{Conclusions and Future Work}

We present a multimodal dialogue system for Conversational Image Editing.  We derived the POMDP formulation for Conversational Image Editing, and our simulated evaluation results show that the DQN policy significantly outperforms a sophisticated rule-based baseline under high semantic error rates.  Our real user study shows that the language understanding component is crucial to success and real users may exhibit more complex behavior.  Future work includes frame-based state tracking, expanding the ontology to incorporate more intents and modeling multimodal user behavior.

\section*{Acknowledgement}
This work is in part supported through collaborative participation in the Robotics Consortium sponsored by the U.S Army Research Laboratory under the Collaborative Technology Alliance Program, Cooperative Agreement W911NF-10-2-0016. This work should not be interpreted as representing the official policies, either expressed or implied, of the Army Research Laboratory of the U.S. Government. The U.S. Government is authorized to reproduce and distribute reprints for Government purposes notwithstanding any copyright notation herein.  We would like to thank the anonymous reviewers for their insightful comments.

\bibliographystyle{plainnat}
\bibliography{nips_2018}

\begin{thebibliography}{22}
\providecommand{\natexlab}[1]{#1}
\providecommand{\url}[1]{\texttt{#1}}
\expandafter\ifx\csname urlstyle\endcsname\relax
  \providecommand{\doi}[1]{doi: #1}\else
  \providecommand{\doi}{doi: \begingroup \urlstyle{rm}\Url}\fi

\bibitem[Anderson et~al.(2018)Anderson, Wu, Teney, Bruce, Johnson,
  S{\"u}nderhauf, Reid, Gould, and van~den Hengel]{anderson2018vision}
Peter Anderson, Qi~Wu, Damien Teney, Jake Bruce, Mark Johnson, Niko
  S{\"u}nderhauf, Ian Reid, Stephen Gould, and Anton van~den Hengel.
\newblock Vision-and-language navigation: Interpreting visually-grounded
  navigation instructions in real environments.
\newblock In \emph{Proceedings of the IEEE Conference on Computer Vision and
  Pattern Recognition (CVPR)}, volume~2, 2018.

\bibitem[Bradski(2000)]{opencv_library}
G.~Bradski.
\newblock {The OpenCV Library}.
\newblock \emph{Dr. Dobb's Journal of Software Tools}, 2000.

\bibitem[Budzianowski et~al.(2017)Budzianowski, Ultes, hao Su, Mrksic, Wen,
  Casanueva, Rojas-Barahona, and Gasic]{Budzianowski2017SubdomainMF}
Pawel Budzianowski, Stefan Ultes, Pei hao Su, Nikola Mrksic, Tsung-Hsien Wen,
  I{\~n}igo Casanueva, Lina~Maria Rojas-Barahona, and Milica Gasic.
\newblock Sub-domain modelling for dialogue management with hierarchical
  reinforcement learning.
\newblock In \emph{SIGDIAL Conference}, 2017.

\bibitem[Bui(2006)]{trung2006multimodal}
Trung Bui.
\newblock Multimodal dialogue management-state of the art.
\newblock 2006.

\bibitem[Das et~al.(2017)Das, Kottur, Gupta, Singh, Yadav, Moura, Parikh, and
  Batra]{visdial}
Abhishek Das, Satwik Kottur, Khushi Gupta, Avi Singh, Deshraj Yadav,
  Jos\'e~M.F. Moura, Devi Parikh, and Dhruv Batra.
\newblock {V}isual {D}ialog.
\newblock In \emph{Proceedings of the IEEE Conference on Computer Vision and
  Pattern Recognition (CVPR)}, 2017.

\bibitem[Dhingra et~al.(2017)Dhingra, Li, Li, Gao, Chen, Ahmed, and
  Deng]{dhingra2017towards}
Bhuwan Dhingra, Lihong Li, Xiujun Li, Jianfeng Gao, Yun-Nung Chen, Faisal
  Ahmed, and Li~Deng.
\newblock Towards end-to-end reinforcement learning of dialogue agents for
  information access.
\newblock In \emph{Proceedings of the 55th Annual Meeting of the Association
  for Computational Linguistics (Volume 1: Long Papers)}, volume~1, pages
  484--495, 2017.

\bibitem[El~Asri et~al.(2017)El~Asri, Schulz, Sharma, Zumer, Harris, Fine,
  Mehrotra, and Suleman]{el2017frames}
Layla El~Asri, Hannes Schulz, Shikhar Sharma, Jeremie Zumer, Justin Harris,
  Emery Fine, Rahul Mehrotra, and Kaheer Suleman.
\newblock Frames: a corpus for adding memory to goal-oriented dialogue systems.
\newblock In \emph{Proceedings of the 18th Annual SIGdial Meeting on Discourse
  and Dialogue}, pages 207--219, 2017.

\bibitem[Hakkani-T{\"u}r et~al.(2014)Hakkani-T{\"u}r, Slaney, Celikyilmaz, and
  Heck]{hakkani2014eye}
Dilek Hakkani-T{\"u}r, Malcolm Slaney, Asli Celikyilmaz, and Larry Heck.
\newblock Eye gaze for spoken language understanding in multi-modal
  conversational interactions.
\newblock In \emph{Proceedings of the 16th International Conference on
  Multimodal Interaction}, pages 263--266. ACM, 2014.

\bibitem[Heck et~al.(2013)Heck, Hakkani-T{\"u}r, Chinthakunta, Tur, Iyer,
  Parthasarathy, Stifelman, Shriberg, and Fidler]{heck2013multi}
Larry Heck, Dilek Hakkani-T{\"u}r, Madhu Chinthakunta, Gokhan Tur, Rukmini
  Iyer, Partha Parthasarathy, Lisa Stifelman, Elizabeth Shriberg, and Ashley
  Fidler.
\newblock Multi-modal conversational search and browse.
\newblock In \emph{First Workshop on Speech, Language and Audio in Multimedia},
  2013.

\bibitem[Kingma and Ba(2014)]{kingma2014adam}
Diederik~P Kingma and Jimmy Ba.
\newblock Adam: A method for stochastic optimization.
\newblock \emph{arXiv preprint arXiv:1412.6980}, 2014.

\bibitem[Laput et~al.(2013)Laput, Dontcheva, Wilensky, Chang, Agarwala, Linder,
  and Adar]{laput2013pixeltone}
Gierad~P Laput, Mira Dontcheva, Gregg Wilensky, Walter Chang, Aseem Agarwala,
  Jason Linder, and Eytan Adar.
\newblock Pixeltone: A multimodal interface for image editing.
\newblock In \emph{Proceedings of the SIGCHI Conference on Human Factors in
  Computing Systems}, pages 2185--2194. ACM, 2013.

\bibitem[Lin et~al.(2014)Lin, Maire, Belongie, Hays, Perona, Ramanan,
  Doll{\'a}r, and Zitnick]{lin2014microsoft}
Tsung-Yi Lin, Michael Maire, Serge Belongie, James Hays, Pietro Perona, Deva
  Ramanan, Piotr Doll{\'a}r, and C~Lawrence Zitnick.
\newblock Microsoft coco: Common objects in context.
\newblock In \emph{European conference on computer vision}, pages 740--755.
  Springer, 2014.

\bibitem[Manuvinakurike et~al.(2018{\natexlab{a}})Manuvinakurike, Brixey, Bui,
  Chang, Artstein, and Georgila]{dial_edit}
Ramesh Manuvinakurike, Jacqueline Brixey, Trung Bui, Walter Chang, Ron
  Artstein, and Kallirroi Georgila.
\newblock {DialEdit}: {Annotations} for {Spoken} {Conversational} {Image}
  {Editing}.
\newblock In \emph{Proceedings of the 14th {Joint} {ACL} - {ISO} {Workshop} on
  {Interoperable} {Semantic} {Annotation}}, Santa Fe, New Mexico, August
  2018{\natexlab{a}}. Association for Computational Linguistics.
\newblock URL \url{https://aclanthology.info/papers/W18-4701/w18-4701}.

\bibitem[Manuvinakurike et~al.(2018{\natexlab{b}})Manuvinakurike, Bui, Chang,
  and Georgila]{manuvinakurike2018conversational}
Ramesh Manuvinakurike, Trung Bui, Walter Chang, and Kallirroi Georgila.
\newblock Conversational image editing: Incremental intent identification in a
  new dialogue task.
\newblock In \emph{Proceedings of the 19th Annual SIGdial Meeting on Discourse
  and Dialogue}, pages 284--295, 2018{\natexlab{b}}.

\bibitem[Manuvinakurike et~al.(2018{\natexlab{c}})Manuvinakurike, Brixey, Bui,
  Chang, Kim, Artstein, and Georgila]{editme}
Ramesh~R. Manuvinakurike, Jacqueline Brixey, Trung Bui, Walter Chang, Doo~Soon
  Kim, Ron Artstein, and Kallirroi Georgila.
\newblock Edit me: {A} corpus and a framework for understanding natural
  language image editing.
\newblock In \emph{{LREC}}. European Language Resources Association {(ELRA)},
  2018{\natexlab{c}}.

\bibitem[Mnih et~al.(2015)Mnih, Kavukcuoglu, Silver, Rusu, Veness, Bellemare,
  Graves, Riedmiller, Fidjeland, Ostrovski, et~al.]{mnih2015human}
Volodymyr Mnih, Koray Kavukcuoglu, David Silver, Andrei~A Rusu, Joel Veness,
  Marc~G Bellemare, Alex Graves, Martin Riedmiller, Andreas~K Fidjeland, Georg
  Ostrovski, et~al.
\newblock Human-level control through deep reinforcement learning.
\newblock \emph{Nature}, 518\penalty0 (7540):\penalty0 529, 2015.

\bibitem[Peng et~al.(2017)Peng, Li, Li, Gao, Çelikyilmaz, Lee, and
  Wong]{Peng2017CompositeTD}
Baolin Peng, Xiujun Li, Lihong Li, Jianfeng Gao, Asli Çelikyilmaz, Sungjin
  Lee, and Kam-Fai Wong.
\newblock Composite task-completion dialogue policy learning via hierarchical
  deep reinforcement learning.
\newblock In \emph{EMNLP}, 2017.

\bibitem[Schatzmann et~al.(2007)Schatzmann, Thomson, Weilhammer, Ye, and
  Young]{schatzmann2007agenda}
Jost Schatzmann, Blaise Thomson, Karl Weilhammer, Hui Ye, and Steve Young.
\newblock Agenda-based user simulation for bootstrapping a pomdp dialogue
  system.
\newblock In \emph{Human Language Technologies 2007: The Conference of the
  North American Chapter of the Association for Computational Linguistics;
  Companion Volume, Short Papers}, pages 149--152. Association for
  Computational Linguistics, 2007.

\bibitem[Ultes et~al.(2017)Ultes, Rojas~Barahona, Su, Vandyke, Kim, Casanueva,
  Budzianowski, Mrk\v{s}i\'{c}, Wen, Gasic, and Young]{ultes2017pydial}
Stefan Ultes, Lina~M. Rojas~Barahona, Pei-Hao Su, David Vandyke, Dongho Kim,
  I\~{n}igo Casanueva, Pawe{\l} Budzianowski, Nikola Mrk\v{s}i\'{c},
  Tsung-Hsien Wen, Milica Gasic, and Steve Young.
\newblock {PyDial: A Multi-domain Statistical Dialogue System Toolkit}.
\newblock In \emph{Proceedings of ACL 2017, System Demonstrations}, pages
  73--78, Vancouver, Canada, July 2017. Association for Computational
  Linguistics.
\newblock URL \url{http://aclweb.org/anthology/P17-4013}.

\bibitem[Wen et~al.(2017)Wen, Vandyke, Mrk{\v{s}}i{\'c}, Gasic, Barahona, Su,
  Ultes, and Young]{wen2017network}
Tsung-Hsien Wen, David Vandyke, Nikola Mrk{\v{s}}i{\'c}, Milica Gasic, Lina
  M~Rojas Barahona, Pei-Hao Su, Stefan Ultes, and Steve Young.
\newblock A network-based end-to-end trainable task-oriented dialogue system.
\newblock In \emph{Proceedings of the 15th Conference of the European Chapter
  of the Association for Computational Linguistics: Volume 1, Long Papers},
  volume~1, pages 438--449, 2017.

\bibitem[Whittaker and Walker(2005)]{whittaker2005evaluating}
Steve Whittaker and Marilyn Walker.
\newblock Evaluating dialogue strategies in multimodal dialogue systems.
\newblock In \emph{Spoken Multimodal Human-Computer Dialogue in Mobile
  Environments}, pages 247--268. Springer, 2005.

\bibitem[Young et~al.(2013)Young, Ga{\v{s}}i{\'c}, Thomson, and
  Williams]{young2013pomdp}
Steve Young, Milica Ga{\v{s}}i{\'c}, Blaise Thomson, and Jason~D Williams.
\newblock Pomdp-based statistical spoken dialog systems: A review.
\newblock \emph{Proceedings of the IEEE}, 101\penalty0 (5):\penalty0
  1160--1179, 2013.

\end{thebibliography}

\appendix

\section{System Interface}\label{sec:interface}

\begin{figure}[H]
    \centering
    \includegraphics[width=\textwidth]{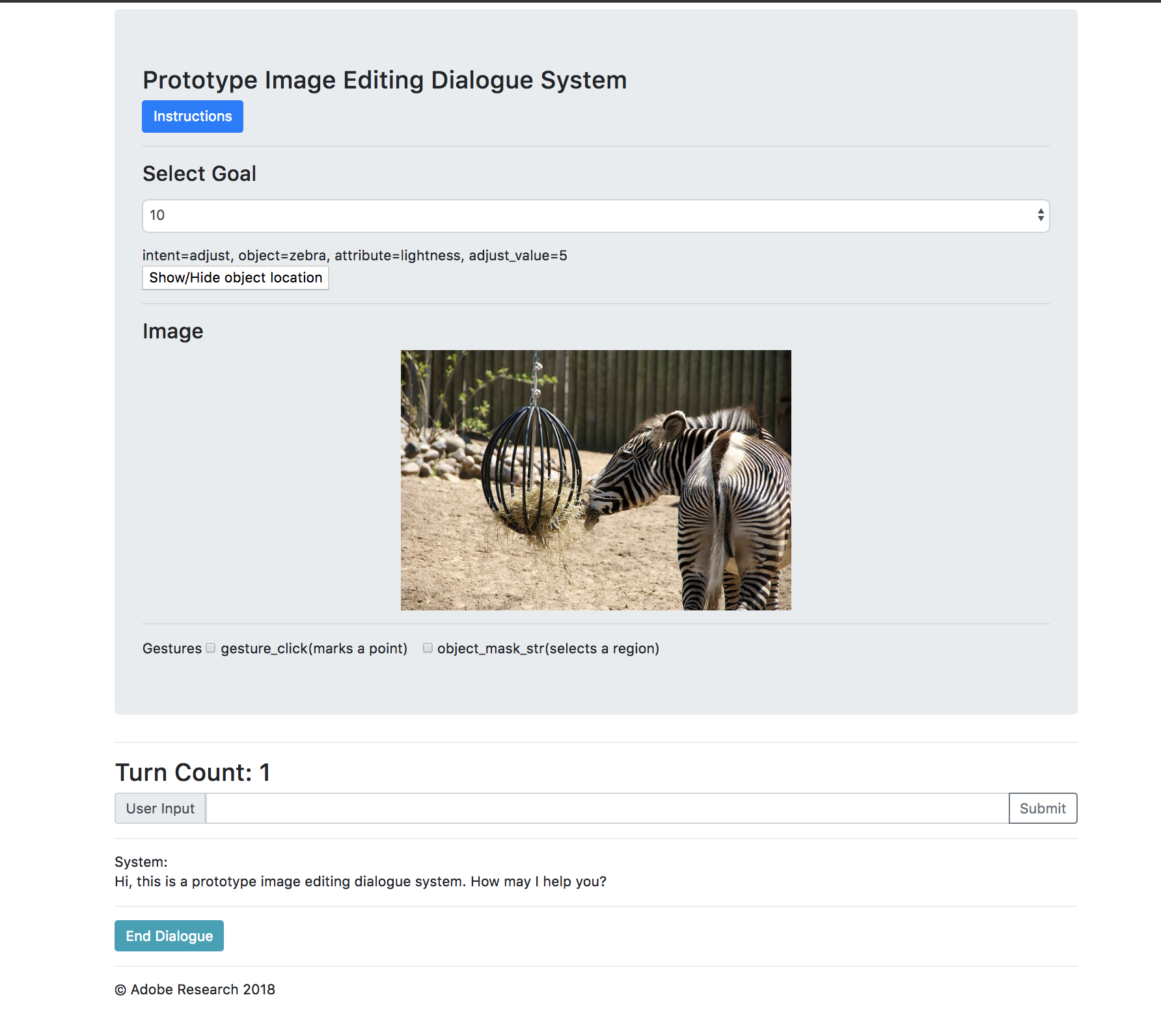}
    \caption{Interface of our system prototype.  A goal is presented to the user in semantic form. Users can input a click (gesture\_click) or select a bounding box (object\_mask\_str)}
    \label{fig:ui}
\end{figure}

\end{document}